\documentclass{article}
\usepackage{spconf}
\usepackage[utf8]{inputenc}
\usepackage{graphicx}
\usepackage[ruled,vlined, algo2e]{algorithm2e}
\usepackage{framed,multirow}
\usepackage{multirow}
\usepackage{amsmath}
\usepackage{blkarray}
\usepackage{comment}
\usepackage{rotating}
\makeatletter
\newcommand\notsotiny{\@setfontsize\notsotiny{8.31415}{9.1828}}
\makeatother

\usepackage{amssymb, bm}
\name{Kateryna Chumachenko$^{\dagger}$\thanks{This  work  was  supported  by  the  Business Finland  project 5G VIIMA.}, Jenni Raitoharju$^{\dagger \mathsection}$, Moncef Gabbouj$^{\dagger}$, Alexandros Iosifidis$^{\star}$   }
\address{$^{\dagger}$Tampere University, Faculty of Information Technology and Communication Sciences, Finland  \\
$^{\mathsection}$ Finnish Environment Institute, Programme for Environmental Information, Finland \\
    $^{\star}$Aarhus University, Department of Engineering,  Denmark 
    }

\begin{document}
\title{Incremental Fast Subclass Discriminant Analysis}
\maketitle

\begin{abstract}
This paper proposes an incremental solution to Fast Subclass Discriminant Analysis (fastSDA). We present an exact and an approximate linear solution, along with an approximate kernelized variant. Extensive experiments on eight image datasets with different incremental batch sizes show the superiority of the proposed approach in terms of training time and accuracy being equal or close to fastSDA solution and outperforming other methods.
\end{abstract}
\section{INTRODUCTION}

Dictated by the high availability of data nowadays, dimensionality reduction has become an essential part of most machine learning applications. Dimensionality reduction refers to the process of finding a mapping from the original space to a subspace of data while satisfying certain statistical criteria. Subspace learning is one of the dominant approaches for this problem and is an essential tool for processing high-dimensional data, such as images or videos \cite{ iosifidis2012multidimensional}. Notable approaches in this area include Linear Discriminant Analysis (LDA) \cite{lda, iosifidis2014kernel, iosifidis2013optimal} and Subclass Discriminant Analysis (SDA) \cite{sdaa}. SDA relaxes the limitations of LDA that are related to unimodality assumption on data, hence becoming a method of choice for applications where data of each class is represented by multimodal distributions, e.g., images taken under different lighting and different cameras.

High-dimensional large-scale data generally suffers from speed limitations when statistical learning approaches are used. A step towards relaxing such limitations of LDA was taken by introducing fast Subclass Discriminant Analysis (fastSDA) \cite{sdaext} that reformulates the original SDA problem and provides a significantly faster solution, making it more applicable to high-dimensional and large-scale data.

In real-world applications, more training data generally becomes available over the time that the model is utilized. In order to take advantage of the newly-acquired data, the model needs to be re-trained from scratch using full data. This process is computationally intensive thus becoming a significant limitation especially for visual data, such as images and videos, which are typically large-scale and relatively high-dimensional. In an ideal case one would like to update the already existing model with newly-available data without repeating the training from scratch. The methods that allow such updates are referred to as incremental learning methods - the set of original data is referred to as the initial batch and further update sets are referred to as incremental batches or update batches.
 
 In this paper, we propose an exact and an approximate incremental solution for the linear fastSDA relying on the utilization of Woodbury identity and an approximate incremental solution for the kernelized formulation of the algorithm relying on incremental Cholesky decomposition \cite{matrixcook, matrixcomp}. Unlike previous solutions for incremental Subclass Discriminant Analysis, the proposed solution is applicable to both scenarios where only one sample is added and where a batch of multiple samples is added. Besides, our solution results in improved accuracy and faster speed as shown by the experiments performed on eight image datasets.

\section{Related Work}
Subspace learning methods aim at finding such subspace of the original data that would satisfy a certain statistical criterion for the data projected onto this subspace, while reducing its dimensionality. Differences between various subspace learning methods generally lie in the definitions of these criteria \cite{ iosifidis2013representative}. LDA is one of the classical supervised methods for dimensionality reduction and its solution is obtained by optimizing the Fisher-Rao's criterion defined over between-class and within-class scatter matrices \cite{fischerraoo}.

\subsection{Fast Subclass Discriminant Analysis}
A step towards relaxing  the limitations related to unimodality assumptions of LDA and extending the possible dimensionality of the learnt subspace has been taken by Subclass Discriminant Analysis \cite{sdaa}. SDA expresses each class with a set of subclasses that are obtained by applying some clustering algorithm on the data of each class. Thus, the between-class and within-class scatter matrices are formulated based on subclass distances rather than class distances. 

Although relaxing some of the limitations, SDA suffers from low speed in high-dimensional and large datasets (in linear and kernelized formulations, respectively). To address this issue, a speed-up approach was recently proposed along with its kernelized form based on the observations of the form of the between-class Laplacian matrix and utilization of Spectral Regression \cite{sdaext}. These methods are referred to as fast Subclass Discriminant Analysis (fastSDA) and fast Kernel Subclass Discriminant Analysis (fastKSDA) for the linear and non-linear cases, respectively. The fastSDA/fastKSDA algorithm can be formulated as follows: 1. Create the between-class Laplacian matrix \cite{sdaext}; 2. Generate the target vectors $\mathbf{t}$ following \cite{sdaext} and create the matrix $\mathbf{T}$ out of the obtained vectors; 3. Regress $\mathbf{T}$ to $\mathbf{W}$ as $\mathbf{W} = (\mathbf{XX}^T + \delta \mathbf{I})^{-1}\mathbf{XT}^T$; 4. Orthogonalize $\mathbf{W}$ such that $\mathbf{W}^T\mathbf{W} = \mathbf{I}$.
Equivalently, for the kernel case, the steps 3-4 are the regression of $\mathbf{T}$ to $\mathbf{A}$: 
$\mathbf{A} = (\mathbf{KK}^T + \delta \mathbf{I})^{-1}\mathbf{KT}^T$,
and orthogonalization of $\mathbf{A}$ such that $\mathbf{A}^T\mathbf{KA} = \mathbf{I}$.

\section{Incremental Fast Subclass Discriminant Analysis}
In this section, we formulate incremental solutions for the fastSDA. Let us denote by $\mathbf{X}_t$, $\mathbf{X}_{t+1}$, and $\mathbf{X} = [\mathbf{X}_t, \mathbf{X}_{t+1}]$ the uncentered data of initial batch, incremental batch, and  total data of both batches, respectively, by $\tilde{\mathbf{X}}_t$ the data of initial batch data centered to own mean, and by $\hat{\mathbf{X}}_t$, $\hat{\mathbf{X}}_{t+1}$, and $\hat{\mathbf{X}}$ the corresponding data centered to the mean of $\mathbf{X}$. 

\subsection{Linear case}
The speed-up of the incremental step is achieved by avoiding the computationally intensive calculation of the $(\hat{\mathbf{X}}\hat{\mathbf{X}}^T)^{-1}$ but updating it based on $(\tilde{\mathbf{X}}_{t}\tilde{\mathbf{X}}_{t}^T)^{-1}$.
The proposed incremental solution is achieved by the following steps: first, the inverse corresponding to the initial batch is re-centered at the mean of total data. Further, the obtained inverse is updated with the data of a new batch.
Given the inverse of the total scatter of the initial batch $(\tilde{\mathbf{X}}_{t}\tilde{\mathbf{X}}_{t}^{T})^{-1}$, we first update it so that it is centered at the mean of the new data. Based on Woodbury identity \cite{matrixcook, matrixcomp}:
{\small 
\begin{eqnarray}
    (\hat{\mathbf{X}}_{t}\hat{\mathbf{X}}_{t}^T)^{-1} &= & (\tilde{\mathbf{X}}_{t}\tilde{\mathbf{X}}_{t}^T + N_{t}\bm{\mu}_{\Delta}\bm{\mu}_{\Delta}^T)^{-1}  \nonumber \\ 
   &=& (\tilde{\mathbf{X}}_{t}\tilde{\mathbf{X}}_{t}^T)^{-1} - (\tilde{\mathbf{X}}_{t}\tilde{\mathbf{X}}_{t}^T)^{-1}\bm{\mu}_{\Delta}\Big(\frac{1}{N_t}
   + \nonumber \\ 
   &+& \bm{\mu}_{\Delta}^T(\tilde{\mathbf{X}}_{t}\tilde{\mathbf{X}}_{t}^T)^{-1}\bm{\mu}_{\Delta}\Big)^{-1}\bm{\mu}_{\Delta}^T(\tilde{\mathbf{X}}_t\tilde{\mathbf{X}}_t^T)^{-1},
\end{eqnarray}}where $N_t$ is the number of samples in initial batch, {\small $\bm{\mu}_{\Delta} = \bm{\mu}_t - \bm{\mu}$}; {\small$\bm{\mu}$} and {\small$\bm{\mu}_{t}$} are the means of {\small$\mathbf{X}$} and {\small$\mathbf{X}_t$}, respectively.

After re-centering the inverse of the scatter of initial batch, we update it with the data of the new batch similarly.
{\small \setlength\arraycolsep{0.1em}
\begin{eqnarray}
    (\hat{\mathbf{X}}\hat{\mathbf{X}}^T)^{-1}& = &
    (\hat{\mathbf{X}}_t\hat{\mathbf{X}}_t^T + \hat{\mathbf{X}}_{t+1}\hat{\mathbf{X}}_{t+1}^{T})^{-1} \nonumber\\  &=& (\hat{\mathbf{X}}_t\hat{\mathbf{X}}_t^T)^{-1} - (\hat{\mathbf{X}}_t\hat{\mathbf{X}}_t^T)^{-1} \hat{\mathbf{X}}_{t+1}\Big(\mathbf{I}\ + \nonumber \\    &+& \hat{\mathbf{X}}_{t+1}^T(\hat{\mathbf{X}}_t\hat{\mathbf{X}}_t^T)^{-1}\hat{\mathbf{X}}_{t+1}\Big)^{-1}\hat{\mathbf{X}}_{t+1}^T(\hat{\mathbf{X}}_t\hat{\mathbf{X}}_t^T)^{-1},
\end{eqnarray}}
where $\mathbf{I}$ is the identity matrix.

Further, to obtain the updated projection matrix $\mathbf{W}'$ the regression step of fastSDA with the updated inverse matrix is applied: $\mathbf{W}' = (\hat{\mathbf{X}}\hat{\mathbf{X}}^T)^{-1}\hat{\mathbf{X}}\mathbf{T}'^T$ and $\mathbf{W}'$ is normalized. In case of regularization, $\mathbf{W}' = (\hat{\mathbf{X}}\hat{\mathbf{X}}^T + \delta\mathbf{I})^{-1}\hat{\mathbf{X}}\mathbf{T}'^T$ can be obtained by using the same regularization parameter $\delta$ as for inverse corresponding to initial batch and substituting ${\tilde{\mathbf{X}}}_t\tilde{\mathbf{X}}_t^T$ by ${\tilde{\mathbf{X}}}_t\tilde{\mathbf{X}}_t^T + \delta\mathbf{I}$ and $\hat{\mathbf{X}}_t\hat{\mathbf{X}}_t^T$ by $\hat{\mathbf{X}}_t\hat{\mathbf{X}}_t^T + \delta\mathbf{I}$ in (1) and (2). Recalculating $\mathbf{T}'$ yields the exact same solution as the original fastSDA solution. An approximate solution can be obtained by updating the original $\mathbf{T}$. The update is achieved as follows: for each class and subclass in the incremental batch, the values of $\mathbf{T}$ corresponding to the same classes/subclasses are replicated to the corresponding positions in $\mathbf{T}'$ and normalizing such that $\mathbf{T}'\mathbf{T}'^{T} = \mathbf{I}$. This can result in further speed-up of the method, while the accuracy is generally close to the one of the exact solution as illustrated by our experiments. 

\subsection{Linear update rule without initial batch}
The proposed linear incremental rule utilizes the data of the initial batch. Often in real-world applications the initial batch data is  
infeasible to store due to e.g. privacy restrictions, loss, or unavailabity for some reasons. Here we extend the previously-described approach in a way that would not utilize the initial data in its update rule, but rely on values calculated during the initial step. Besides the above-mentioned limitations, this allows to process data that is too large for performing certain calculations, allowing to solve the problem by incrementally updating the projection matrix. 

For simplicity of notation, let {\small$(\tilde{\mathbf{X}}_{t}\tilde{\mathbf{X}}_{t}^T)^{-1} = a$, \linebreak $(\tilde{\mathbf{X}}_{t}\tilde{\mathbf{X}}_{t}^T)^{-1}\tilde{\mathbf{X}}_{t} = A$; $(\hat{\mathbf{X}}_{t}\hat{\mathbf{X}}_{t}^T)^{-1} = b$, $(\hat{\mathbf{X}}_{t}\hat{\mathbf{X}}_{t}^T)^{-1}\hat{\mathbf{X}}_{t} = B$; $(\hat{\mathbf{X}}\hat{\mathbf{X}}^T)^{-1} = c$, $(\hat{\mathbf{X}}\hat{\mathbf{X}}^T)^{-1}\hat{\mathbf{X}} = C.$} Thus, we will obtain $C$ by utilizing $A$ and $a$ by first centering $A$ at the new mean, and further updating it with new data. Following (1), $b$ can be obtained as {\small $b = a - a\bm{\mu}_{\Delta}(\frac{1}{N_t}
+\bm{\mu}_{\Delta}^Ta\bm{\mu}_{\Delta})^{-1}\bm{\mu}_{\Delta}^Ta$}. The re-centering of $A$ can be achieved following \cite{matrixcook, matrixcomp} as follows:
{\small 
\begin{eqnarray}
    B &=& (\hat{\mathbf{X}}_{t}\hat{\mathbf{X}}_{t}^T)^{-1}\hat{\mathbf{X}}_{t} =  (\tilde{\mathbf{X}}_{t}\tilde{\mathbf{X}}_{t}^T + N_{t}\bm{\mu}_{\Delta}\bm{\mu}_{\Delta}^T)^{-1}(\tilde{\mathbf{X}}_{t} + \bm{\mu}_{\Delta}\mathbf{1}_{N_{t}}^T) \nonumber\\ \nonumber 
&=& b(\tilde{\mathbf{X}}_{t} + \bm{\mu}_{\Delta}\mathbf{1}_{N_{t}}^T) = A - a\bm{\mu}_{\Delta} \Big(\frac{1}{N_t}
    + \bm{\mu}_{\Delta}^Ta\bm{\mu}_{\Delta}\Big)^{-1}\bm{\mu}_{\Delta}^TA + \\ &+& a\bm{\mu}_{\Delta}\mathbf{1}_{N_{t}}^T - a\bm{\mu}_{\Delta}\Big(\frac{1}{N_t}
    + \bm{\mu}_{\Delta}^Ta\bm{\mu}_{\Delta}\Big)^{-1} \bm{\mu}_{\Delta}^Ta\bm{\mu}_{\Delta}\mathbf{1}_{N_{t}}^T,
\end{eqnarray}}where $\mathbf{1}_{N_{t}}$ is a vector of ones.

After centering $A$ at the new mean, the result can be updated with the new data following (2):
{\small 
\begin{eqnarray}
     c &=& 
    b - b\hat{\mathbf{X}}_{t+1}(\mathbf{I}+\hat{\mathbf{X}}_{t+1}^Tb\hat{\mathbf{X}}_{t+1})^{-1}\hat{\mathbf{X}}_{t+1}^Tb.
\end{eqnarray}}Finally, $C$ can be obtained as follows \cite{matrixcook, matrixcomp}: 
{\small 
\begin{eqnarray}
        C &=& (\hat{\mathbf{X}}\hat{\mathbf{X}}^T)^{-1}\hat{\mathbf{X}} 
        = [C1 \hspace{0.2cm} C2 ], \nonumber\\ 
        C1 &=& (\hat{\mathbf{X}}\hat{\mathbf{X}}^T)^{-1}\hat{\mathbf{X}}_t = B - c\hat{\mathbf{X}}_{t+1}\hat{\mathbf{X}}_{t+1}^TB, \nonumber \\ 
        C2 &=& (\hat{\mathbf{X}}\hat{\mathbf{X}}^T)^{-1}\hat{\mathbf{X}}_{t+1} =  c\hat{\mathbf{X}}_{t+1}.
\end{eqnarray}}

\subsection{Nonlinear update rule}
For obtaining the non-linear solution incrementally, we note that the solution of the kernel regression problem for orthogonalized $\mathbf{A}$ is given by solving 
{\small
\begin{equation}
    \mathbf{A} = (\mathbf{K} + \delta\mathbf{I})^{-1}\mathbf{T}^T,
\end{equation}}which can be solved using Cholesky decomposition:
{\small
\begin{equation}
    \mathbf{R} = chol(\mathbf{K} + \delta\mathbf{I}); \:\:\:
    \mathbf{A} = (\mathbf{R}^{-1})^T\mathbf{R}^{-1}\mathbf{T}^T.
\end{equation} }We also note that {\small$\mathbf{K}_{t+1} = \left[\begin{array}{cc}
    \mathbf{K}_t &  \mathbf{K}_{t,t+1}\\
     \mathbf{K}_{t+1,t} & \mathbf{K}_{t+1,t+1} 
\end{array}\right]$.} Therefore, the new inverse can be obtained by updating the upper-triangular matrix of the Cholesky decomposition $\mathbf{R}$ as follows \cite{incrchol, cskda}:
{\small
\begin{equation}\mathbf{R}_{t+1} = \left[\begin{array}{cc}
    \mathbf{R}_t &  \mathbf{R}_{t,t+1}\\
     \mathbf{0} & \mathbf{R}_{t+1,t+1} 
\end{array}\right],
\end{equation}}
{\small
\begin{equation}
\mathbf{R}_{t+1,t+1} = chol(\mathbf{K}_{t+1,t+1} - \mathbf{R}_{t,t+1}^T\mathbf{R}_{t,t+1} + \delta\mathbf{I})
\end{equation}}
{\small
\begin{equation}
    \mathbf{R}_{t,t+1} = \mathbf{R}_{t}^{-T}\mathbf{K}_{t,t+1}
\end{equation}}
\:\:\:\:\:\:Note that in this case, re-centering the data in $\mathcal{F}$ yields a different matrix $\mathbf{K}$, resulting in inapplicability of incremental Cholesky decomposition. Therefore, we re-center the incremental batches of the data at the mean of the initial batch, hence, the obtained solution is approximate. 
After obtaining the updated inverse, the regression step of (6) is applied followed by the normalization of $\mathbf{A}'$. The target vector matrix can be updated similarly to the linear case or recalculated. The speed-up given by this solution is two-fold: first, the speed-up is achieved as the inverse calculation is omitted; second, the resulting method does not require recalculation of the full kernel matrix, unlike kernel fastSDA and KSDA, as they rely on mean-centering of the data in $\mathcal{F}$.

\begin{table*}[h!]
\centering
\setlength{\tabcolsep}{2pt}
\footnotesize
\caption{Classification   results   of   linear   methods:   accuracy/number   of   clusters   per   class/time   in   sec.}
\begin{tabular}{|l|l|c|c|c|c|c|c|c|c|c|}   
\hline
                                                         &   \textbf{Dataset}      &   \textbf{fastSDA}                                                &  \textbf{IA-fastSDA}                                       &   \textbf{I-fastSDA}                                          &   \textbf{IA-fastSDA-NB}                                 &   \textbf{I-fastSDA-NB}                                 &   \textbf{LDA}                              &   \textbf{I-LDA}                           &   \textbf{SDA}                                                            &   \textbf{I-SDA}                              \\   
\hline
\multirow{11}{*}{\rotatebox[origin=c]{90}{\textit{\hspace{0.8cm}1 Sample}}}   &   BU                                                &   \textbf{64.9/1~ ~0.334}                                 &   63.6/1~ ~0.017                                                            &   \textbf{64.9/1~ ~0.017}                                 &   63.6/1~ ~0.042                                                            &   \textbf{64.9\textbf{/1}~ ~0.037}      &   62.1~ ~2.423                                    &    -                                                                  &   62.1/1~ ~0.502                                                            &   -                                                                     \\
                                                         &   SEMEION                                 &   87.5/4~ ~0.016                                                            &   86.7/4~ ~0.002                                                            &   87.5/4~ ~0.003                                                            &   86.7/4~ ~0.009                                                            &   87.5/4~  ~0.006                                                                  &   88.7~ ~0.154                                    &   -                                                                  &   \textbf{91.0/4~  ~0.052}                                       &   -                                                                     \\
                                                         &   YALE                                          &   {84.0/1~     ~0.442}                                                                  &   84.5/1~ ~0.035                                                            &   84.0/1~ ~0.041                                          &   84.5/1~ ~0.095                                                            &   84.0/1~ ~0.099                                                                  &   \textbf{86.7~ ~7.363}         &   -                                                                  &   \textbf{86.7\textbf{/1}~ ~3.289}      &   -                                                                     \\
                                                         &   CAL7-CENT                           &   \textbf{91.9\textbf{/1}~ ~0.014}      &   91.7/1~ ~0.003                                                            &   \textbf{91.9\textbf{/1}~ ~0.003}      &   91.7/1~ ~0.008                                                            &   \textbf{91.9\textbf{/1}~ ~0.007}      &   90.0~ ~0.103                                          &   -                                                                  &   90.6/3~ ~0.026                                                            &    -                                                                     \\
                                                         &   CAL20-CENT                        &   \textbf{77.3\textbf{/1}~ ~0.016}      &   77.2/1~ ~0.003                                                            &   \textbf{77.3\textbf{/1}~ ~0.005}      &   77.2/1~ ~0.011                                                            &   \textbf{77.3\textbf{/1}~ ~0.010}         &   76.1~ ~0.080&   -                                                                  &   77.0/1~ ~0.026                                                                  &    -                                                                     \\
                                                         &   CAL20-HOG                           &   89.3/2~ ~2.202                                                            &   89.5/2~ ~0.099                                                            &   89.3/2~ ~0.095                                                            &   89.5/2~ ~0.199                                                            &   89.3/2~ ~0.208                                                                  &   \textbf{92.8~ ~17.26}                                                         &   -                                                                  &   \textbf{92.8/1~ ~4.664}                                 &    -                                                                     \\
                                                         &   CARDAMAGE                                    &   \textbf{92.8/1~ ~2.810}                                    &   \textbf{92.8/1~ ~0.098}                                 &   \textbf{92.8/1~ ~0.094}                                 &   \textbf{92.8/1~ ~0.211}                                    &   \textbf{92.8/1~ ~0.190}                                    &   91.7~ ~15.60                                                        &   -                                                                  &   92.5/2~ ~3.046                                                            &    -                                                                     \\
                                                         &   LANDUSE                                             &   \textbf{95.9/2~ ~2.743}                                 &   \textbf{95.9/1~ ~0.084}                                 &   \textbf{95.9/2~ ~0.091}                                 &   \textbf{95.9/1~ ~0.192}                                 &   \textbf{95.9/2~ ~0.182}                                 &   95.7~ ~16.80                                                         &   -                                                                  &   95.7/1~ ~5.591                                                            &   -                                                                     \\   
\hline
\multirow{11}{*}{\rotatebox[origin=c]{90}{\textit{\hspace{0.8cm}Batch 10\%}}}   &   BU                                                &   \textbf{62.6/1~ ~0.355}                                 &   62.1/1~ ~0.019                                                            &   \textbf{62.6/1~ ~0.021}                                    &   62.1/1~ ~0.049                                                            &   \textbf{62.6/1~ ~0.044}                                 &   62.1~ ~2.592                                    &   27.6~ ~1.057                                       &   62.1/1~ ~0.523                                                            &   38.7/4~ ~1.766                                    \\
                                                         &   SEMEION                                 &   87.5/2~ ~0.014                                                            &   87.6/3~ ~0.005                                                            &   87.5/2~ ~0.005                                                            &   87.6/3~ ~0.009                                                            &   87.5/2~ ~0.008                                                            &   \textbf{88.7~ ~0.058}         &   56.8~ ~0.033                                       &   \textbf{90.3/2~ ~0.024}                                 &   79.8/2~ ~0.029                                    \\
                                                         &   YALE                                          &   93.7/1~ ~0.440                                                               &   93.8/1~ ~0.057                                                            &   93.7/1~ ~0.056                                                            &   \textbf{93.8\textbf{/1}~ ~0.108}      &   93.7/1~ ~0.109                                                            &   86.7~ ~5.987                                    &   80.1~ ~1.870                                          &   86.7/1~ ~3.258                                                            &   84.0/2~ ~11.34                                       \\
                                                         &   CAL7-CENT                           &   \textbf{91.9\textbf{/1}~ ~0.013}      &   \textbf{91.9\textbf{/1}~ ~0.003}      &   \textbf{91.9\textbf{/1}~ ~0.004}      &   \textbf{91.9\textbf{/1}~ ~0.007}      &   \textbf{91.9\textbf{/1}~ ~0.007}      &   90.0~ ~0.076                                          &   55.4~ ~0.026                                       &   90.4/2~ ~0.016                                                            &   71.5/4~ ~0.033                                    \\
                                                         &   CAL20-CENT                        &   77.3/1~ ~0.015      &   \textbf{77.5\textbf{/1}~ ~0.006}      &   77.3/1~ ~0.007                                 &   \textbf{77.5\textbf{/1}~ ~0.013}      &   77.3/1~ ~0.011                                                            &   76.1~ ~0.090                                                         &   69.8~ ~0.078                                                            &   77.0/1~ ~0.026                                                                  &   59.7\textbf{/1}~ ~0.024         \\
                                                         &   CAL20-HOG                           &   91.7/1~ ~2.193                                                            &   91.8/1~ ~0.124                                                            &   91.7/1~ ~0.122                                                            &   91.8/1~ ~0.228                                                            &   91.7/1~ ~0.222                                                            &   \textbf{92.8~ ~19.93}      &   65.8~ ~6.203      &   \textbf{92.8\textbf{/1}~ ~4.652}      &   81.5\textbf{/1}~ ~4.570            \\
                                                         &   CARDAMAGE                                    &   \textbf{92.8\textbf{/1}~ ~2.768}      &   \textbf{92.8\textbf{/1}~ ~0.123}      &   \textbf{92.8\textbf{/1}~ ~0.135}      &   \textbf{92.8\textbf{/1}~ ~0.228}      &   \textbf{92.8\textbf{/1}~ ~0.238}      &   91.8~ ~18.64                                 &   82.7~ ~7.031                                       &   92.5/2~ ~2.964                                                            &   84.4/2~ ~2.682                                    \\
                                                         &   LANDUSE                                             &   95.9/3~ ~2.914                                                            &   \textbf{96.2/3~ ~0.157}                                 &   95.9/3~ ~0.121                                                            &   \textbf{96.2/3~ ~0.272}                                 &   95.9/3~ ~0.225                                                            &   95.7~ ~18.10                                 &   74.0~ ~7.725                                             &   95.7/1~ ~6.437                                                            &   78.4/2~ ~13.15                                    \\   
\hline
\multirow{11}{*}{\rotatebox[origin=c]{90}{\textit{\hspace{0.8cm}Batch 30\%}}}   &   BU                                                &   61.7/2~ ~0.651                                                            &   61.9/2~ ~0.068                                                            &   61.7/2~ ~0.071                                                            &   61.9/2~ ~0.112                                                            &   61.7/2~ ~0.112                                                            &   \textbf{62.1~ ~5.434}         &   28.0~ ~2.990                                                            &   \textbf{62.1/1~ ~0.954}                                 &   41.0/5~ ~5.062                                          \\
                                                         &   SEMEION                                 &   86.4/3~ ~0.043                                                            &   85.8/4~ ~0.057                                                            &   86.4/3~ ~0.068                                                            &   85.8/3~ ~0.062                                                            &   86.4/4~ ~0.074                                                            &   88.7~ ~0.154                                    &   56.2~ ~0.112                                       &   \textbf{90.8/4~ ~0.141}                                 &   79.6/2~ ~0.072                                    \\
                                                         &   YALE                                          &   \textbf{94.0/1~ ~0.892}                                       &   93.8/1~ ~0.297                                                            &   \textbf{94.0/1~ ~0.325}                                       &   93.8/1~ ~0.407                                                            &   \textbf{94.0/1~ ~0.399}                                       &   86.7~ ~13.03                                 &   79.9~ ~7.065                                       &   86.7/1~ ~6.765                                                            &   83.8/5~ ~71.90                                 \\
                                                         &   CAL7-CENT                           &   \textbf{91.9\textbf{/1}~ ~0.012}      &   \textbf{91.9\textbf{/1}~ ~0.011}      &   \textbf{91.9\textbf{/1}~ ~0.013}      &   \textbf{91.9\textbf{/1}~ ~0.015}      &   \textbf{91.9\textbf{/1}~ ~0.015}      &   89.9~ ~0.085                                                         &   55.6~ ~0.034                                       &   90.6/4~ ~0.034                                                            &   71.1/2~ ~0.019                                    \\
                                                         &   CAL20-CENT                        &   \textbf{77.3/1~ ~0.015}                                 &   77.2/1~ ~0.028                                                            &   \textbf{77.3\textbf{/1}~ ~0.030}         &   77.2/1~ ~0.037                                                            &   \textbf{77.3\textbf{/1}~ ~0.032}      &   76.1~ ~0.102                                                            &   65.3~ ~0.081                                                            &   77.0/1~ ~0.025                                                                  &   59.8/1~ ~0.027                                    \\
                                                         &   CAL20-HOG                           &   91.0/1~ ~2.196                                                                  &   91.3/1~ ~0.170                                                               &   91.0/1~ ~0.181                                                                  &   91.3/1~ ~0.283                                                            &   91.0/1~ ~0.285                                                                  &   \textbf{92.8~ ~19.62}      &   64.9~ ~9.553                                       &   \textbf{92.8\textbf{/1}~ ~4.531}      &   80.8/1~ ~4.516                                    \\
                                                         &   CARDAMAGE                                    &   92.4/2~ ~2.871                                                            &   \textbf{92.5/2~ ~0.235}                                 &   92.4/2~ ~0.235                                                            &   \textbf{92.5/2~ ~0.322}                                 &   92.4/2~ ~0.313                                                            &   91.8~ ~18.97                                 &   82.7~ ~11.27                                    &   92.3/4~ ~3.208                                                            &   82.8/1~ ~2.739                                    \\
                                                         &   LANDUSE                                             &   95.7/1~ ~2.698                                                            &   \textbf{96.1/1~ ~0.202}                                 &   95.7/1~ ~0.206                                                            &   \textbf{96.1/1~ ~0.306}                                 &   95.7/1~ ~0.302                                                            &   95.7~ ~18.03                                 &   73.9~ ~11.48                                    &   95.7/1~ ~5.439                                                            &   80.6/2~ ~13.18                                 \\
\hline
\end{tabular}
\setlength{\tabcolsep}{4.2pt}
\footnotesize
\begin{tabular}{|l|l|c|c|c|c|c|c|c|c|c|}
\multicolumn{10}{c}{ }\\
\multicolumn{10}{c}{\normalsize{\textbf{Table 2.} Classification results of kernel methods: accuracy/number of clusters per class/time in sec.}}\\
\hline
\multicolumn{6}{|c}{ Centered } & \multicolumn{4}{|c|}{ Non-centered }\\
\hline                             &\textbf{Datasets} & \textbf{fastKSDA}                                       & \textbf{IA-fastKSDA}                                    & \textbf{I-fastKSDA}                                     & \textbf{KSDA}                                           & \textbf{fastKSDA}                                       & \textbf{IA-fastKSDA}                                    & \textbf{I-fastKSDA}                                     & \textbf{KSDA}                                            \\
                             \hline
\multirow{8}{*}{\rotatebox[origin=c]{90}{\textit{\hspace{0.1cm}1 Sample}}}    & BU                & 62.4/1~ ~0.015          & 63.3/1~ ~0.006          & 63.1/1~ ~0.006          & 61.6/2~ ~0.059          & 63.3/2~ ~0.011          & 62.0/1~ ~0.004            & \textbf{63.9/1~ ~0.005} & 58.0/2~ ~0.034           \\
                             & SEMEION           & 95.1/1~ ~0.074          & 95.0/1~ ~0.021            & 95.0/1~ ~0.022            & 95.3/1~ ~2.181          & 95.0/1~ ~0.023            & \textbf{95.5/1~ ~0.009} & 95.2/1~ ~0.010           & 95.0/1~ ~1.372             \\
                             & YALE              & 84.1/3~ ~0.234          & 88.1/3~ ~0.070           & 88.1/3~ ~0.073          & \textbf{91.3/1~ ~0.728} & 78.1/1~ ~0.074          & 77.4/1~ ~0.029          & 77.5/2~ ~0.036          & 85.6/5~ ~0.696           \\
                             & CAL7-CENT         & 92.4/1~ ~0.071          & 91.7/1~ ~0.022          & 91.7/1~ ~0.022          & \textbf{92.5/4~ ~1.201} & 92.1/1~ ~0.024          & 91.8/1~ ~0.010           & 92.1/1~ ~0.011          & 89.6/1~ ~1.911           \\
                             & CAL20-CENT        & \textbf{80.4/1~ ~0.214} & 80.0/1~ ~0.067            & 80.0/1~ ~0.068            & 78.2/1~ ~4.379          & 80.1/1~ ~0.054          & 80.2/1~ ~0.023          & 80.1/1~ ~0.025          & 75.8/4~ ~2.750            \\
                             & CAL20-HOG         & 87.3/1~ ~0.270           & 86.5/1~ ~0.076          & 86.5/1~ ~0.077          & 90.9/1~ ~8.382          & 87.1/1~ ~0.102          & 87.7/1~ ~0.037          & 86.9/1~ ~0.039          & \textbf{90.7/1~ ~9.062}  \\
                             & CARDAMAGE         & 93.2/1~ ~0.217          & 93.2/1~ ~0.059          & 93.2/1~ ~0.060           & 93.2/1~ ~7.290          & \textbf{93.3/1~ ~0.099} & \textbf{93.3/1~ ~0.099} & \textbf{93.3/1~ ~0.030}  & 76.3/5~ ~5.749           \\
                             & LANDUSE           & 96.8/1~ ~0.183          & 96.8/1~ ~0.054          & 96.8/1~ ~0.056          & 96.1/1~ ~3.353          & 96.7/1~ ~0.080           & 96.7/1~ ~0.029          & \textbf{96.9/1~ ~0.034} & 94.7/2~ ~3.184           \\
                             \hline
\multirow{8}{*}{\rotatebox[origin=c]{90}{\textit{\hspace{0.1cm} Batch 10\%}}} & BU                & 62.4/1~ ~0.016          & 62.9/1~ ~0.006          & 63.3/1~ ~0.007          & 60.5/1~ ~0.034          & 63.0/1~ ~0.011            & 62.3/1~ ~0.006          & \textbf{63.9/1~ ~0.006} & 55.0/1~ ~0.068             \\
                             & SEMEION           & 95.1/1~ ~0.073          & 95.0/1~ ~0.022            & 95.1/1~ ~0.024          & 94.8/3~ ~0.215          & 95.0/1~ ~0.024            & \textbf{95.5/1~ ~0.012} & 95.2/2~ ~0.013          & 94.5/1~ ~1.432           \\
                             & YALE              & 83.1/1~ ~0.225          & 81.0/5~ ~0.097            & 80.8/5~ ~0.102          & \textbf{88.9/1~ ~0.738} & 78.1/1~ ~0.076          & 77.3/1~ ~0.044          & 77.2/1~ ~0.046          & 83.7/4~ ~0.700             \\
                             & CAL7-CENT         & \textbf{92.5/1~ ~0.066} & 91.6/1~ ~0.022          & 91.5/1~ ~0.023          & 91.6/2~ ~1.303          & 92.1/1~ ~0.021          & 91.9/1~ ~0.011          & 92.1/1~ ~0.012          & 90.5/1~ ~0.172           \\
                             & CAL20-CENT        & 80.3/1~ ~0.218          & \textbf{80.5/1~ ~0.073} & \textbf{80.5/1~ ~0.074} & 78.9/1~ ~0.765          & 80.1/1~ ~0.054          & 80.3/1~ ~0.032          & 80.1/1~ ~0.034          & 79.0/1~ ~0.689           \\
                             & CAL20-HOG         & 87.3/1~ ~0.271          & 86.5/1~ ~0.096          & 86.5/1~ ~0.097          & 89.7/1~ ~7.058          & 87.1/1~ ~0.109          & 87.7/1~ ~0.052          & 86.9/1~ ~0.055          & \textbf{91.1/1~ ~8.948}  \\
                             & CARDAMAGE         & 93.2/1~ ~0.206          & 93.2/1~ ~0.081          & 93.2/1~ ~0.080           & \textbf{93.3/1~ ~7.053} & \textbf{93.3/1~ ~0.094} & \textbf{93.3/1~ ~0.047} & \textbf{93.3/1~ ~0.048} & 74.7/5~ ~7.170            \\
                             & LANDUSE           & 96.8/1~ ~0.207          & 96.9/1~ ~0.073          & \textbf{97.0/1~ ~0.076}   & 96.8/1~ ~2.734          & 96.7/1~ ~0.094          & 96.7/1~ ~0.050           & 96.9/1~ ~0.051          & 93.9/2~ ~1.364           \\
                             \hline
\multirow{8}{*}{\rotatebox[origin=c]{90}{\textit{\hspace{0.1cm}Batch 30\%}}}  & BU                & 62.6/1~ ~0.038          & 63.4/1~ ~0.020           & 59.0/2~ ~0.023            & 58.9/2~ ~0.145          & 63.0/1~ ~0.025            & 62.7/1~ ~0.018          & \textbf{63.7/1~ ~0.019} & 57.9/1~ ~0.065           \\
                             & SEMEION           & 95.1/1~ ~0.184          & 95.0/1~ ~0.089            & 95.1/1~ ~0.091          & \textbf{95.4/1~ ~4.391} & 95.0/1~ ~0.052            & 95.3/1~ ~0.050           & 95.2/1~ ~0.054          & 94.8/1~ ~2.854           \\
                             & YALE              & \textbf{89.9/1~ ~0.561} & 75.9/3~ ~0.141          & 76.9/3~ ~0.145          & 89.5/1~ ~1.652          & 78.1/1~ ~0.173          & 77.0/1~ ~0.161            & 77.4/2~ ~0.147          & 84.2/4~ ~0.827           \\
                             & CAL7-CENT         & \textbf{92.5/1~ ~0.073} & 91.7/1~ ~0.031          & 90.7/2~ ~0.030           & 91.6/1~ ~1.061          & 92.1/1~ ~0.022          & 91.9/1~ ~0.017          & 92.1/1~ ~0.018          & 88.7/3~ ~1.202           \\
                             & CAL20-CENT        & 80.2/1~ ~0.209          & \textbf{80.3/1~ ~0.098} & 80.0/1~ ~0.099            & 80.0/1~ ~0.771            & 80.1/1~ ~0.053          & \textbf{80.3/1~ ~0.045} & 80.1/1~ ~0.046          & 75.7/4~ ~4.162           \\
                             & CAL20-HOG         & 87.3/1~ ~0.263          & 86.7/1~ ~0.131          & 86.6/1~ ~0.133          & \textbf{90.3/1~ ~6.920}  & 87.1/1~ ~0.107          & 87.9/1~ ~0.079          & 86.9/1~ ~0.081          & 88.4/3~ ~4.737           \\
                             & CARDAMAGE         & 93.2/1~ ~0.214          & 93.3/1~ ~0.111          & \textbf{93.3/1~ ~0.111} & \textbf{93.3/1~ ~7.335} & \textbf{93.3/1~ ~0.095} & \textbf{93.3/1~ ~0.077} & \textbf{93.3/1~ ~0.079} & 71.0/4~ ~7.244           \\
                             & LANDUSE           & 96.8/1~ ~0.188          & 96.8/1~ ~0.095          & 96.8/1~ ~0.096          & \textbf{96.9/1~ ~3.492} & 96.7/1~ ~0.081          & 96.7/1~ ~0.069          & \textbf{96.9/1~ ~0.071} & 95.3/1~ ~2.266   \\
                             \hline
\end{tabular}
\end{table*}
\section{Experiments}
We compare the results of the proposed algorithms on eight image datasets for different tasks. For the face recognition task, we utilize the Extended YALE-B dataset \cite{YALE}, containing facial images of 38 individuals taken under different lighting conditions, resulting in 2432 grayscale images. Another facial image dataset is the BU dataset \cite{BU} consisting of 700 samples. The dataset is used for facial expression recognition thus defining a classification problem with 7 classes. 
In both facial image datasets, each sample is rescaled to a $30\times40$ image and further flattened to obtain a 1200-dimensional vector. For the application of digit recognition, we use Semeion dataset containing 1593 samples of handwritten digits represented by 16x16 binarized images flattened to vectors of length 256 \cite{semeion}. For the object recognition task, two subsets of Caltech-101 dataset of 7 and 20 classes are considered, consisting of 1474 and 2386 instances, respectively, \cite{caltech, gitdata}. CENTRIST features of length 254 are extracted from both datasets and 1984-dimensional HOG features are additionaly considered for Caltech-101-20 dataset. UC Merced Land Use Dataset \cite{landuse} consists of 2100 samples and contains aerial orthoimagery samples of 21 different land use types, including harbor, river, forest, and freeway. The final task considered in this paper uses a subset of 2399 samples from the Car Damage Detection dataset \cite{cardamage} containing images of damaged and non-damaged cars. 2048-dimensional features extracted from the pre-last layer of ResNet-50 \cite{resnet50} pre-trained on ImageNet were utilized for the latter two datasets.

We compare the proposed approaches with SDA \cite{sdaa}, fastSDA \cite{sdaext}, incremental LDA (I-LDA) \cite{incrlda}, and incremental SDA (I-SDA) \cite{incrsda}. We compare the time required by the incremental step of the incremental methods with the time of recalculation of the projection matrix for the full data of both batches for non-incremental methods. Subclass labels for the initial batch are obtained by clustering the original data using $k$-means clustering. Subclass labels for the incremental batch are obtained using the subclass centroids of the initial batch for all the methods to eliminate the effect of clustering on accuracy, and clustering time is not included in total training time. In kernel variants, training time includes the time taken for kernel matrix calculation. We use an RBF kernel with $\sigma$ set to the mean distance between the training vectors. In incremental methods, $\sigma$ calculated on the initial batch is used for the incremental batch. 
We also compare the exact solution obtained by recreating the target vectors and the approximate solution obtained by updating them. For the kernel case, we report the results for data mean-centered at the mean of the initial batch and the non-centered data. Classification is achieved by $kNN$ classifier with $k=5$. To assess the scalability of the solution, multiple incremental batch sizes are evaluated: 30\%, 10\% of training data and 1 sample, with the rest of training data being the initial batch. Incremental SDA and Incremental LDA rely on the calculation of the scatter matrices of incremental batch data, hence they are not applicable for the 1-sample update case.
We perform 5-fold stratified cross-validation where 50\% of data is used for training, 30\% - for validation, and 20\% - for testing. The validation set is used for fine-tuning the regularization parameters, and the results are reported by training on the training set and testing on the test set. The regularization parameter is chosen from the set \{$10^{-3},10^{-2},10^{-1},1,10,100,1000$\}. Dimensionality of the projected space is determined by the rank of the between-class scatter matrix and is set to $Cz-1$ for all subclass-based methods, and $C$-1 for LDA and their incremental variants, where $C$ is the number of classes and $z$ is the number of subclasses per class. The number of subclasses from 1-5 were evaluated and the best result is reported.

The results for linear and kernelized methods are shown in Tables 1 and 2. We report the accuracy along with the number of clusters per class and training time in seconds. IA-fastSDA and I-fastSDA refer to the proposed incremental approach with the update of target vector matrix (i.e., approximate method) and its recreation (i.e., exact method), respectively. Similarly, IA-fastSDA-NB and I-fastSDA-NB refer to those methods which are not utilizing the initial batch data. As can be observed, the proposed incremental solutions result in faster speed than the initial fastSDA and I-SDA/LDA in the majority of the cases. The results obtained with the recalculation of the target vectors are exactly equal to those of fastSDA in terms of accuracy. In the kernelized formulation, non-centering generally results in a better performance.
\section{Conclusion}
Exact and approximate incremental solutions for fastSDA along with the kernel variants are proposed in this paper. The methods were evaluated on eight image datasets and were shown to be efficient and effective, making them a reasonable choice for incremental learning on multimodal high-dimensional large-scale data.


\pagebreak
\bibliography{bibliography.bib}

\begin{thebibliography}{10}

\bibitem{semeion}
M.~Buscema.
\newblock Metanet: the theory of independent judges.
\newblock {\em Substance Use \& Misuse}, 33:439--461, 1998.

\bibitem{sdaext}
K.~Chumachenko, J.~Raitoharju, A.~Iosifidis, and M.~Gabbouj.
\newblock Speed-up and multi-view extensions to subclass discriminant analysis.
\newblock {\em arXiv preprint arXiv:1905.00794}, 2019.

\bibitem{caltech}
L.~Fei-Fei, R.~Fergus, and P.~Perona.
\newblock One-shot learning of object categories.
\newblock {\em IEEE Transactions on Pattern Recognition and Machine
  Intelligence}, 8:594--611, 2006.

\bibitem{fischerraoo}
R.~Fisher.
\newblock The statistical utilization of multiple measurements.
\newblock {\em Annals of Eugenics}, 8:376--386, 1938.

\bibitem{matrixcomp}
G.~Golub and C.~van Loan.
\newblock {\em Matrix computations}, volume~3.
\newblock JHU press, 2012.

\bibitem{resnet50}
K.~He, X.~Zhang, S.~Ren, and J.~Sun.
\newblock Deep residual learning for image recognition.
\newblock In {\em Proceedings of the IEEE Conference on Computer Vision and
  Pattern Recognition}, pages 770--778, 2016.

\bibitem{cskda}
A.~Iosifidis and M.~Gabbouj.
\newblock Class-specific kernel discriminant analysis revisited: further
  analysis and extensions.
\newblock {\em IEEE Transactions on Cybernetics}, 47:4485--4496, 2017.

\bibitem{iosifidis2012multidimensional}
A.~Iosifidis, A.~Tefas, and I.~Pitas.
\newblock Multidimensional sequence classification based on fuzzy distances and
  discriminant analysis.
\newblock {\em IEEE Transactions on Knowledge and Data Engineering},
  25(11):2564--2575, 2012.

\bibitem{iosifidis2013optimal}
A.~Iosifidis, A.~Tefas, and I.~Pitas.
\newblock On the optimal class representation in linear discriminant analysis.
\newblock {\em IEEE transactions on neural networks and learning systems},
  24(9):1491--1497, 2013.

\bibitem{iosifidis2013representative}
A.~Iosifidis, A.~Tefas, and I.~Pitas.
\newblock Representative class vector clustering-based discriminant analysis.
\newblock In {\em 2013 Ninth International Conference on Intelligent
  Information Hiding and Multimedia Signal Processing}, pages 526--529. IEEE,
  2013.

\bibitem{iosifidis2014kernel}
A.~Iosifidis, A.~Tefas, and I.~Pitas.
\newblock Kernel reference discriminant analysis.
\newblock {\em Pattern Recognition Letters}, 49:85--91, 2014.

\bibitem{incrlda}
T.~Kim, S.~Wong, B.~Stenger, J.~Kittler, and R.~Cipolla.
\newblock Incremental linear discriminant analysis using sufficient spanning
  set approximations.
\newblock In {\em 2007 IEEE Conference on Computer Vision and Pattern
  Recognition}, pages 1--8. IEEE, 2007.

\bibitem{incrsda}
H.~Lamba, T.~Dhamecha, M.~Vatsa, and R.~Singh.
\newblock Incremental subclass discriminant analysis: A case study in face
  recognition.
\newblock In {\em 2012 19th IEEE International Conference on Image Processing},
  pages 593--596. IEEE, 2012.

\bibitem{YALE}
K.~Lee, J.~Ho, and D.~Kriegman.
\newblock Acquiring linear subspaces for face recognition under variable
  lightning.
\newblock {\em IEEE Transactions on Pattern Analysis and Machine Intelligence},
  27:684--698, 2005.

\bibitem{gitdata}
Y.~Li, F.~Nie, H.~Huang, and J.~Huang.
\newblock Large-scale multi-view spectral clustering via bipartite graph.
\newblock In {\em AAAI Conference on Artificial Intelligence}, pages
  2750--2756, 2015.

\bibitem{matrixcook}
K.~Petersen, M.~Pedersen, et~al.
\newblock The matrix cookbook, vol. 7.
\newblock {\em Technical University of Denmark}, 15, 2008.

\bibitem{cardamage}
A.~Shah.
\newblock Car damage detection: Damaged and whole cars image dataset, 2019.
\newblock Retrieved December 2019 from
  https://www.kaggle.com/anujms/car-damage-detection.

\bibitem{incrchol}
G.~Stewart.
\newblock {\em Matrix Algorithms: Volume 1: Basic Decompositions}, volume~1.
\newblock Siam, 1998.

\bibitem{landuse}
Y.~Yang and S.~Newsam.
\newblock Bag-of-visual-words and spatial extensions for land-use
  classification.
\newblock In {\em Proceedings of the 18th SIGSPATIAL International Conference
  on Advances in Geographic Information Systems}, pages 270--279, 2010.

\bibitem{lda}
J.~Ye.
\newblock Least squares linear discriminant analysis.
\newblock {\em International Conference on Machine Learning}, 1:1087--1093,
  2007.

\bibitem{BU}
L.~Yin, X.~Wei, Y.~Sun, J.~Wang, and M.~Rosato.
\newblock A 3d facial expression database for facial behavior research.
\newblock In {\em IEEE International Conference on Automatic Face and Gesture
  Recognition}, pages 211--216, Southampton, UK, 2006.

\bibitem{sdaa}
M.~Zhu and A.~Martinez.
\newblock Subclass discriminant analysis.
\newblock {\em IEEE Transactions on Pattern Analysis and Machine Intelligence},
  28, 2006.

\end{thebibliography}
\end{document}